\documentclass{IEEEtaes}

\usepackage{amsmath,amssymb}
\usepackage{graphicx}
\usepackage{cite}
\usepackage{booktabs}
\usepackage{multirow}
\usepackage{url}

\jvol{XX}
\jnum{XX}
\jmonth{XXXXX}
\paper{XXXXXXX}
\pubyear{2026}
\doiinfo{TAES.2026.Doi Number}
\begin{document}

\title{Neural Operator Learning for Collision-Aware Trajectory Planning of Spacecraft Swarms}

\author{Sidhdharth D. Sikka}
\affil{Purdue University, West Lafayette, IN, USA}

\author{Suyi Gao}
\affil{Purdue University, West Lafayette, IN, USA}

\author{Zehui Lu}
\affil{Independent Researcher, Fremont, CA, USA}

\author{Rongjie Lai}
\affil{Purdue University, West Lafayette, IN, USA}

\author{Shaoshuai Mou}
\affil{Purdue University, West Lafayette, IN, USA}

\receiveddate{This work has been submitted to the IEEE for possible publication. Copyright may be transferred without notice, after which this version may no longer be accessible.}

\corresp{{\itshape (Corresponding author: Sidhdharth D. Sikka)}}

\authoraddress{Sidhdharth D. Sikka is with Manifold Research Group and the School of Aeronautics and Astronautics, Purdue University, West Lafayette, IN 47907 USA (e-mail: sikkas@purdue.edu). Suyi Gao and Rongjie Lai are with the Department of Mathematics, Purdue University, West Lafayette, IN 47907 USA (e-mail: gao757@purdue.edu; lairj@purdue.edu). Zehui Lu is an independent researcher, Fremont, CA USA (e-mail: zehuilu789@gmail.com). Shaoshuai Mou is with the School of Aeronautics and Astronautics, Purdue University, West Lafayette, IN 47907 USA (e-mail: mous@purdue.edu).}

\markboth{SIKKA ET AL.}{NEURAL OPERATOR LEARNING FOR COLLISION-AWARE TRAJECTORY PLANNING OF SPACECRAFT SWARMS}
\maketitle

\begin{abstract}
Satellite constellations require orbital transfers that are both fuel efficient and collision avoidant. Yet, the computational cost of optimization methods traditionally used to plan their trajectories scales poorly with both the number of satellites as well as the number of obstacles to avoid, due to the pairwise safety constraints. In this work, we introduce a permutation-equivariant neural operator for trajectory planning of spacecraft swarms. This neural operator maps distributions of spacecraft initial states, target states, and obstacle initial states to trajectories which avoid collision and conserve fuel. This neural operator output is then paired with a batched Gauss--Newton finish to enforce exact orbital dynamics, and further reduce fuel use. The operator is self-supervised, trained without optimal trajectory labels. When trained on ten spacecraft, the proposed method generalized zero-shot to swarms of 1,000 spacecraft and 11,000 obstacles. The generated trajectories matched a per-agent optimal control solver's accuracy while retaining collision avoidance. Operator learning grounded in physics may offer a fast, scalable alternative to trajectory optimization in the increasingly crowded orbits of the future.
\end{abstract}


\section{Introduction}\label{sec:intro}

Low Earth Orbit (LEO) is becoming increasingly crowded as satellite constellations expand and debris accumulates. Roughly 16{,}000 active satellites and an estimated 140 million debris fragments now occupy near-Earth space \cite{esa2026environment}. Between December 2025 and May 2026 alone, SpaceX's Starlink constellation executed more than 207{,}000 automated collision-avoidance maneuvers, over three times its rate a year earlier \cite{spacecom2026starlink}. Trajectory planning is thus becoming a persistent, large-scale coordination problem; frequent avoidance maneuvers reduce mission lifetime, consume fuel, and create coordination burdens that grow with constellation size.

Classical approaches to multi-agent orbital maneuvering rely primarily on centralized optimization: mixed-integer linear programs offer safety guarantees \cite{lee2023regional}, linearized, reachable-set, and distributed convex reformulations improve tractability \cite{basu2023collision,cui2024reachable}, and model predictive and receding-horizon schemes handle constraints in closed loop \cite{eren2017mpc,chen2024trajectory}, but all must re-solve programs whose collision constraints multiply with the number of agents and debris objects. Reactive schemes such as velocity obstacles \cite{vandenberg2011orca}, artificial potential fields \cite{pedari2023novel}, energy-constrained formation feedback \cite{babazadeh2020distance}, and deployed rule-based screening avoid this cost but grow conservative and prone to mutual conflict in dense traffic \cite{spacecom2026starlink}, while heuristic \cite{jung2024genetic}, reinforcement-learning \cite{xu2024reinforcement,qu2022proximity,meng2025obstacle}, meta-learning \cite{li2021meta}, and decision-theoretic \cite{kuhl2025mdp} planners, including learned swarm-navigation policies \cite{an2026swarmnav}, are flexible but seldom transfer across swarm sizes and debris densities.

\begin{figure}
    \centering
    \includegraphics[width=\linewidth]{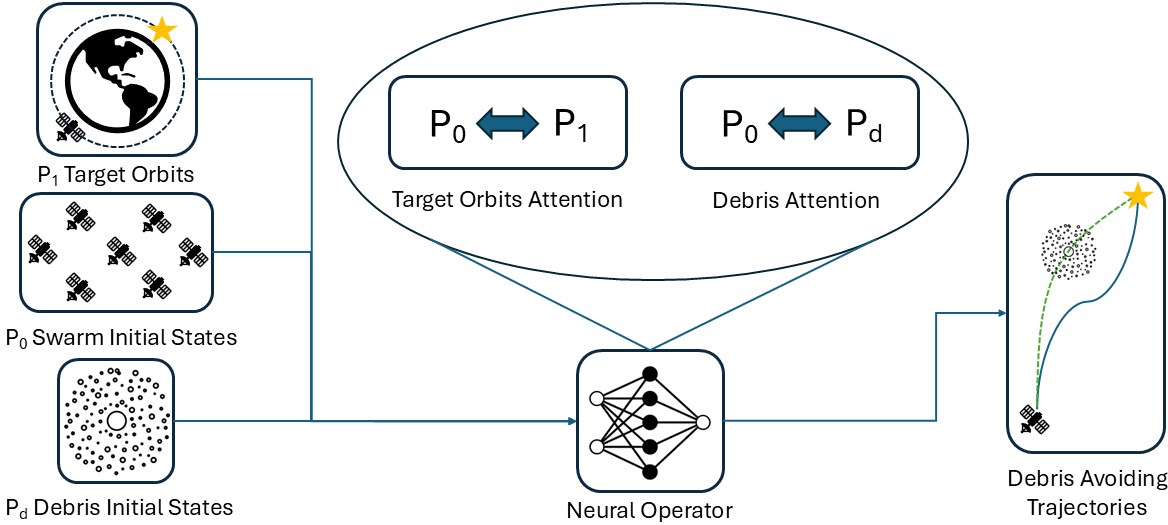}
    \caption{Neural operator framework for collision-aware swarm planning. Spacecraft, target, and debris sets are encoded as distributions; dual attention pathways condition a physics-informed baseline (green dashed) plus learned residual (blue solid), and a batched Gauss--Newton finish closes each trajectory onto exact two-body dynamics.}
    \label{fig:summary}
\end{figure}

\textit{A key difficulty is structural.} Collision constraints couple each spacecraft to other spacecraft and debris objects through pairwise interactions whose count grows rapidly with swarm size and debris density. Rather than treating a swarm as a collection of individual spacecraft, one can instead model it as a probability distribution evolving under controlled dynamics. Mean Field Games (MFGs) formalize this perspective by characterizing the limit of infinitely many interacting agents through coupled Hamilton–Jacobi–Bellman and Fokker–Planck equations \cite{bensoussan2013mean}. While this representation scales more naturally with population size, solving the associated partial differential equations remains computationally demanding in high-dimensional physical domains, even with dedicated machine-learning solvers \cite{ruthotto2020mfg,lauriere2022mfgsurvey}.

Learning offers a way to amortize this cost without requiring labeled optimal solutions: physics-informed networks embed governing equations directly in the training objective \cite{raissi2019pinn,karniadakis2021piml}, and differentiable simulation trains control policies end-to-end through the system dynamics \cite{zhang2025agileflight}. Operator learning extends this paradigm to families of problem instances: neural operators learn mappings between functional inputs and outputs, amortizing inference across instances~\cite{lu2021deeponet,berner2026operators,Wangetal2021,Yangetal2023,Xiaoetal2024,xu2025self}. Distribution-driven control methods have demonstrated that differentiable particle simulations can be used to train neural operators that map initial distributions to target configurations without explicitly solving MFG equations \cite{huang2024unsupervised,cole2026context}.

Here we introduce a two-stage planner for collision-aware trajectory planning of spacecraft swarms in dense debris fields: a self-supervised, permutation-equivariant, time-conditioned neural operator, followed by a lightweight per-agent Gauss--Newton finish that closes each predicted trajectory onto exact two-body dynamics. The operator maps distributions of initial orbits, target configurations, and debris, together with the mission duration, to trajectories for the entire swarm in a single forward pass and, because the same interaction rule applies to any number of sample points, generalizes across swarm sizes without retraining. The method is summarized in Fig.~\ref{fig:summary}.

The main contributions of this work are:
\begin{itemize}\setlength\itemsep{1pt}
\item a self-supervised neural operator for swarm trajectory planning, trained from physics-informed objectives (terminal accuracy, fuel-effort surrogates, and closest-point-of-approach penalties) with no optimal-trajectory labels; and
\item a batched Gauss--Newton finish that restores exact Keplerian dynamics to every predicted trajectory at near-constant cost in swarm size.
\end{itemize}

The remainder of this paper is organized as follows. Section~\ref{sec:methods} develops the problem setup, operator architecture, training objectives, and Gauss--Newton finish; Section~\ref{sec:results} presents Monte Carlo evaluations on real ephemeris data; and Section~\ref{sec:discussion} concludes with implications and limitations.

\section{Methodology}\label{sec:methods}

\subsection{Problem setup and notation}\label{sec:methods_setup}

We consider a swarm of $N$ controlled spacecraft operating in the presence of $M$ unpowered debris objects over a fixed horizon $T$. Each spacecraft state is represented in Keplerian orbital elements,
\[
\boldsymbol{x}(t) =
\begin{bmatrix}
a & \varepsilon & \iota & \Omega & \omega & \nu
\end{bmatrix}^\top \in \mathbb{R}^6,
\]
where $a$ is the semi-major axis, $\varepsilon$ is the eccentricity, $\iota$ is the inclination, $\Omega$ is the right ascension of the ascending node, $\omega$ is the argument of periapsis, and $\nu$ is the true anomaly. Each debris object is represented similarly as $\boldsymbol{d}(t)\in\mathbb{R}^6$. Spacecraft apply control accelerations in the Radial--Transverse--Normal (RTN) frame,
\[
\boldsymbol{u}(t)=
\begin{bmatrix}
u_\mathrm{R} & u_\mathrm{S} & u_\mathrm{W}
\end{bmatrix}^\top \in \mathbb{R}^3,
\]
where $u_\mathrm{R}$, $u_\mathrm{S}$, and $u_\mathrm{W}$ denote the radial, along-track, and cross-track acceleration components, respectively.
\subsection{Orbital dynamics}\label{sec:methods_dynamics}

Spacecraft orbital element dynamics are described by the Gauss Variational Equations (GVE),
\begin{subequations}\label{eq:gve}
\begin{align}
&\dot a = \frac{2 a^2}{h}\!\left(\varepsilon\sin(\nu)\,u_\mathrm{R} + \frac{p}{r}\,u_\mathrm{S}\right),\\
&\dot \varepsilon = \frac{1}{h}\!\left(p\sin(\nu)\,u_\mathrm{R} + \big((p+r)\cos(\nu)+ r \varepsilon\big)\,u_\mathrm{S}\right),\\
&\dot \iota = \frac{r\cos(\theta)}{h}\,u_\mathrm{W},\\
&\dot\Omega = \frac{r\sin(\theta)}{h\sin(\iota)}\,u_\mathrm{W},\\
&\dot\omega = \frac{1}{h \varepsilon}\!\left(-p\cos(\nu)\,u_\mathrm{R} + (p+r)\sin(\nu)\,u_\mathrm{S}\right) \notag\\
&\qquad\qquad - \frac{r\sin(\theta)}{h}\cot(\iota)\,u_\mathrm{W},\\
&\dot \nu = \frac{h}{r^2} + \frac{1}{h \varepsilon}\!\left(p\cos(\nu)\,u_\mathrm{R} - (p+r)\sin(\nu)\,u_\mathrm{S}\right),
\end{align}
\end{subequations}
with
\begin{equation}\label{eq:gve_aux}
\begin{gathered}
p = a(1-\varepsilon^2),\qquad
r=\frac{p}{1+\varepsilon\cos(\nu)},\\
h = \sqrt{\mu\,p},\qquad
\theta = \omega + \nu.
\end{gathered}
\end{equation}

For compactness, we also write the controlled dynamics as
\begin{equation}\label{eq:control_affine}
\dot{\boldsymbol{x}} = \boldsymbol{f}(\boldsymbol{x}) + \boldsymbol{g}(\boldsymbol{x})\boldsymbol{u},
\end{equation}
where $\boldsymbol{f}$ captures the uncontrolled evolution and $\boldsymbol{g}$ collects the control-affine coefficients implied by equation~\eqref{eq:gve}.
Debris are modeled as unpowered ($\boldsymbol{u}\equiv\boldsymbol{0}$) and propagated under the corresponding uncontrolled dynamics $\boldsymbol{\dot{d}} = \boldsymbol{f}(\boldsymbol{d})$.

\subsection{Distributional formulation}\label{sec:methods_distribution}

Let $P_0$ denote the initial spacecraft distribution over $\mathbb{R}^6$, $P_1$ the target distribution over $\mathbb{R}^5$ (for the first five elements), and $P_d$ the initial debris distribution over $\mathbb{R}^6$.
We seek a trajectory map $F(\boldsymbol{x},t)$ whose pushforward defines the time-varying swarm distribution,
\begin{equation}
P_t = F(\cdot,t)_\# P_0.
\end{equation}
Rather than solving a large coupled optimal control problem directly for each $(P_0,P_d,P_1,T)$ instance, we learn an operator that amortizes this mapping across scenarios.

\subsection{Neural trajectory operator}\label{sec:methods_operator}

We define a time-conditioned neural operator
\begin{equation}\label{eq:operator_def}
\mathcal{G}_\theta : (P_0, P_d, P_1, t, T) \mapsto F(\cdot,t)_\# P_0,
\end{equation}
implemented as a permutation-equivariant transformer with multi-head attention. Our architecture builds upon the operator-learning framework introduced in Huang et al.~\cite{huang2024unsupervised}, which uses sampling-invariant, permutation-equivariant attention blocks to learn solution operators for mean-field games from samples of the initial and terminal distributions.

\begin{figure*}[!t]
    \centering
    \includegraphics[width=0.85\textwidth]{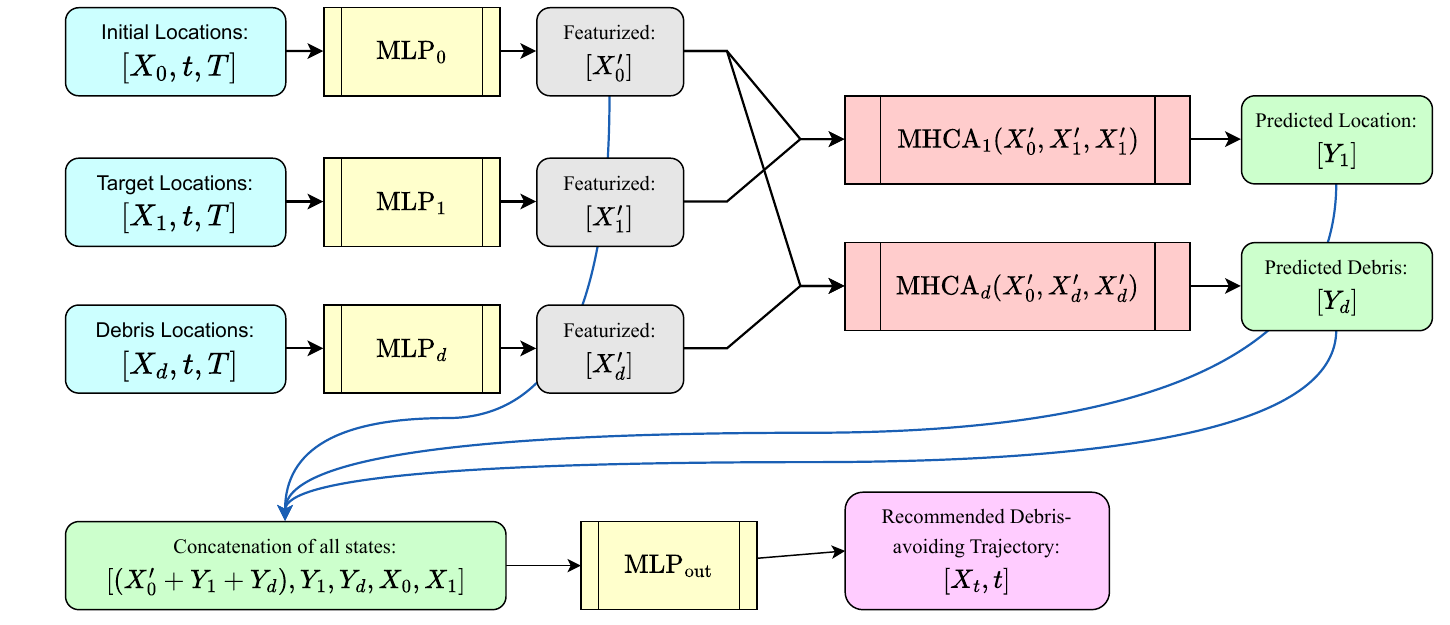}
    \caption{Architecture of the solution operator. $X_0$, $X_1$, and $X_d$ are variable-length sets of spacecraft, target, and debris states; MLP denotes a multilayer perceptron and MHCA multi-head cross-attention.}
    \label{fig:arch}
\end{figure*}

In contrast, our modified network introduces two key differences tailored to the spacecraft-swarm navigation setting: first, we include dual cross-attention streams in parallel, one between agents and the target orbits, and one between agents and the debris set, rather than a unified attention block over all inputs. We maintain a shallow projection head that concatenates fused agent features, target-attention features, debris-attention features, and the original lifted queries, to output per-agent orbital-element predictions. Detail on the composition of the operator is presented in Fig.~\ref{fig:arch}.

\subsection{Terminal loss}\label{sec:methods_terminal}

Given samples $\{\boldsymbol{x}_{0,i}\}_{i=1}^N \sim P_0$ and associated target samples $\{\boldsymbol{x}_{\mathrm{des},i}\}_{i=1}^N \sim P_1$, the operator produces terminal states $\boldsymbol{x}_i(T)=F(\boldsymbol{x}_{0,i},T)$.
We penalize mismatch in the first five elements using
\begin{equation}\label{eq:terminal_loss}
\mathcal{T}\!\left(F(\cdot,T)_\# P_0; P_1\right)
= \frac{1}{N}\sum_{i=1}^{N}
\left\|
\Pi_5\!\big(\boldsymbol{x}_i(T)\big) - \boldsymbol{x}_{\mathrm{des},i}
\right\|_2^2,
\end{equation}
where $\Pi_5(\cdot)$ projects onto $(a,\varepsilon,\iota,\Omega,\omega)$.

\subsection{Interaction penalties}\label{sec:methods_interactions}

Collision avoidance is enforced using a closest-point-of-approach (CPA) penalty evaluated in Cartesian space.
A state $\boldsymbol{x}$ expressed in Keplerian elements is converted to its Earth-centered inertial (ECI) position--velocity state through the standard Keplerian element-to-Cartesian map $\Gamma$, written $\boldsymbol{\chi} = \Gamma(\boldsymbol{x}) = (\boldsymbol{p},\boldsymbol{v})$, and this conversion is applied to every spacecraft and debris state at every timestep.
Debris are propagated as unpowered objects by advancing only the true anomaly $\nu$ under two-body Keplerian motion, keeping $(a,\varepsilon,\iota,\Omega,\omega)$ fixed.

Given a spacecraft state $\boldsymbol{\chi}_{s} = (\boldsymbol{p}_s,\boldsymbol{v}_s)$ and a debris state $\boldsymbol{\chi}_d = (\boldsymbol{p}_d,\boldsymbol{v}_d)$ at the start of a discrete interval of duration $\Delta t$, we compute the relative position and velocity
\[
\Delta\boldsymbol{p} = \boldsymbol{p}_s - \boldsymbol{p}_d,\qquad
\Delta\boldsymbol{v} = \boldsymbol{v}_s - \boldsymbol{v}_d.
\]
The CPA time is
\[
t_{\mathrm{cpa}}(\boldsymbol{\chi}_s, \boldsymbol{\chi}_d) = \mathrm{clip}\!\left(
-\frac{\Delta\boldsymbol{p}^\top \Delta\boldsymbol{v}}{\|\Delta\boldsymbol{v}\|_2^2+\epsilon},
\,0,\,\Delta t\right),
\]
and the CPA separation is
\begin{equation}
\label{eq:cpa_linear_within_interval}
d_{\mathrm{cpa}}(\boldsymbol{\chi}_s, \boldsymbol{\chi}_d)=\left\|\Delta\boldsymbol{p}+\Delta\boldsymbol{v}\,t_{\mathrm{cpa}}(\boldsymbol{\chi}_s, \boldsymbol{\chi}_d)\right\|_2.
\end{equation}
We penalize violations of a pair-type-dependent safety radius using a quadratic hinge, evaluated at every interval along the horizon,
\begin{align}
&\mathcal{J}_{\mathrm{CPA}}(P_s,P_d) = \notag\\
&\mathbb{E}_{\boldsymbol{x}_s\sim P_s,\,\boldsymbol{x}_d\sim P_d}
\left(\kappa\max\!\left\{0,\;r_{d}-d_{\mathrm{cpa}}(\boldsymbol{\chi}_s,\boldsymbol{\chi}_d)\right\}\right)^2 \notag \\
&+w_{s}\,
\mathbb{E}_{\boldsymbol{x}_s,\,\boldsymbol{x}'_s\sim P_s}
\left(\kappa\max\!\left\{0,\;r_{s}-d_{\mathrm{cpa}}(\boldsymbol{\chi}_s,\boldsymbol{\chi}'_s)\right\}\right)^2,
\label{equ:J_cpa_def}
\end{align}
where $P_s$ and $P_d$ are the spacecraft and debris populations, $\boldsymbol{\chi}=\Gamma(\boldsymbol{x})$ is the mapped ECI state of each sampled element vector, and the second expectation excludes the self-pair $\boldsymbol{x}'_s = \boldsymbol{x}_s$.
The spacecraft--debris safety radius $r_{d}=1$~km applies at all times; the spacecraft--spacecraft safety radius $r_{s}=100$~m applies only for $t\geq 0.2\,T$ and its term carries the weight $w_{s}=10$; $\kappa$ is a scaling constant.
The spacecraft--spacecraft radius matches the $100$~m threshold used in evaluation, and the grace window over the first $20\%$ of the transfer exempts the deliberately conflicting clustered start (see Section~\ref{sec:training_setup}), which is unavoidable by construction, while still penalizing any conflict the swarm has not resolved by mid-transfer, including the converging arrival.

\paragraph{Adversarial debris generation.}
To stress-test avoidance behavior during training, every debris object is generated adversarially as a \emph{crossing-orbit} threat to the model's \emph{nominal} rollout, the trajectory produced when the model is prompted with an empty debris set.
Concretely, for a given $(P_0,P_1,T)$, we first perform a rollout with $P_d=\varnothing$ to obtain each agent's nominal trajectory.
For each debris object we select an agent and a random hit time $t_{\mathrm{hit}}$ in the latter half of the transfer, and take the agent's nominal ECI state $(\boldsymbol{p},\boldsymbol{v})$ at $t_{\mathrm{hit}}$.
The debris velocity is the agent's velocity rotated by a random angle drawn from $[20^\circ,75^\circ]$ about a random axis perpendicular to it; the rotation preserves speed, so the debris orbit remains bound, but crosses the agent's path with a substantial relative velocity, as in a genuine conjunction, rather than trailing it co-orbitally.
The debris position is offset from the agent's by a sub-safety-radius near-miss distance ($0.35$--$0.65\,r_{d}$, random direction), the perturbed state is converted back to orbital elements, and, because the debris propagation model advances only $\nu$, the initial true anomaly $\nu_0$ is chosen so that the object reaches this state at $t_{\mathrm{hit}}$ under Keplerian motion.
The near-miss offset is essential: a threat placed at exact position--velocity coincidence produces a closest-approach distance of zero, at which the CPA penalty's gradient with respect to position vanishes identically, so the model receives a large penalty but no direction in which to evade.
The crossing near-miss instead yields a well-conditioned avoidance gradient on every sample.

\subsection{Fuel-cost surrogate from Gauss variational dynamics}\label{sec:methods_fuel_gve}

To encourage fuel-efficient transfers, we penalize the magnitude of the control acceleration implied by the Gauss Variational Equations (GVE). The element-rate dynamics take the control-affine form $\dot{\boldsymbol{x}} = \boldsymbol{f}(\boldsymbol{x}) + \boldsymbol{g}(\boldsymbol{x})\,\boldsymbol{u}$, where $\boldsymbol{x}\in\mathbb{R}^6$ is the orbital-element state, $\boldsymbol{u}\in\mathbb{R}^3$ is the control acceleration in the rotating RTN frame, $\boldsymbol{f}\in\mathbb{R}^6$ is the uncontrolled Keplerian rate (nonzero only in the $\nu$ component), and $\boldsymbol{g}(\boldsymbol{x}) \in \mathbb{R}^{6\times 3}$ collects the GVE control-affine coefficients. Given a predicted trajectory $\boldsymbol{x}(t)$, we estimate $\dot{\boldsymbol{x}}(t)$ by central finite differences over the discretized rollout and infer the corresponding control via the Moore--Penrose pseudoinverse $\boldsymbol{g}^\dagger$:
\[
\hat{\boldsymbol{u}}(t)=\boldsymbol{g}(\boldsymbol{x}(t))^\dagger\left(\dot{\boldsymbol{x}}(t)-\boldsymbol{f}(\boldsymbol{x}(t))\right).
\]
We then define the fuel-cost surrogate as the time integral of squared control magnitude,
\begin{equation}
    \mathcal{F}=\int_0^T \|\hat{\boldsymbol{u}}(t)\|_2^2\,dt,
    \label{equ:fuel_cost_raw}
\end{equation}
implemented in discrete time using the rollout grid.
The term preserves the physical meaning of minimizing RTN control effort without requiring an optimal control solution during training.

\subsection{Trajectory parameterization with a biased baseline and learned residual}\label{sec:methods_bias_rollout}

We represent each spacecraft trajectory in orbital elements as a smooth baseline transfer augmented by a learned residual.
Let $\tau=t/T\in[0,1]$ denote normalized time, and let $\boldsymbol{x}(t)=[a,\varepsilon,\iota,\Omega,\omega,\nu]^\top$.
For the first five ``slow'' elements $(a,\varepsilon,\iota,\Omega,\omega)$, we define a deterministic baseline $\bar{\boldsymbol{x}}_{1:5}(\tau)$ that interpolates between the initial and target values, and we learn an additive bias (residual) $\Delta \boldsymbol{x}_{1:5}(\tau)$:
\[
\boldsymbol{x}_{1:5}(\tau)=\bar{\boldsymbol{x}}_{1:5}(\tau)+\Delta\boldsymbol{x}_{1:5}(\tau).
\]
The baseline uses linear interpolation for $(a,\varepsilon,\iota)$ and wrapped interpolation on the principal branch for angular elements $(\Omega,\omega)$ via
\[
\mathrm{angdiff}(\alpha,\beta)=\mathrm{atan2}(\sin(\alpha-\beta),\cos(\alpha-\beta)),
\]
The resulting slow-element sequence is projected by clamping physical bounds (e.g., $a>0$, $0\le \varepsilon<1$, $0<\iota<\pi$) and wrapping angles to $[0,2\pi)$.

The true anomaly $\nu$ is then propagated forward in time using a physically grounded Keplerian rate with an additional learned correction.
Specifically, at discrete times $t_k$ with steps $\Delta t_k$, we update
\[
\nu_{k+1}=\nu_k+\Delta t_k\Big(\dot{\nu}_{\mathrm{kep}}(a_k,\varepsilon_k,\nu_k)+\Delta \dot{\nu}_k\Big),
\]
where $\dot{\nu}_{\mathrm{kep}}$ is the two-body Keplerian rate and $\Delta\dot{\nu}_k$ is the network output evaluated at the normalized time $\tau_k = t_k / T \in [0,1]$.

\subsection{Training objective}\label{sec:methods_training_obj}

For each sampled scenario $(P_0,P_d,P_1,T)$, the underlying planning task can be viewed as a finite-horizon collision-aware optimal control problem. Given spacecraft samples $\{\boldsymbol{x}_{0,i}\}_{i=1}^N\sim P_0$, target samples $\{\boldsymbol{x}_{\mathrm{des},i}\}_{i=1}^N\sim P_1$, and debris samples $\{\boldsymbol{d}_{0,j}\}_{j=1}^M\sim P_d$, the ideal per-instance problem is to find trajectories and controls that minimize fuel expenditure while reaching the target distribution and avoiding close approaches:
\begin{equation}
\label{eq:instance_ocp}
\begin{aligned}
\min_{\{\boldsymbol{x}_i,\boldsymbol{u}_i\}_{i=1}^N}\quad
& \sum_{i=1}^{N}\int_0^T \|\boldsymbol{u}_i(t)\|_2^2\,dt
+ \lambda_I\,\mathcal{J}_{\mathrm{CPA}} \notag\\
&\quad + \lambda_T \sum_{i=1}^{N}
\big\|
\Pi_5(\boldsymbol{x}_i(T))
-
\Pi_5(\boldsymbol{x}_{\mathrm{des},i})
\big\|_2^2  \\
\mathrm{s.t.}\quad
& \dot{\boldsymbol{x}}_i(t)
= \boldsymbol{f}(\boldsymbol{x}_i(t))
+ \boldsymbol{g}(\boldsymbol{x}_i(t))\boldsymbol{u}_i(t), \\
& \dot{\boldsymbol{d}}_j(t)
= \boldsymbol{f}(\boldsymbol{d}_j(t)), \\
& \boldsymbol{x}_i(0)=\boldsymbol{x}_{0,i},
\quad
\boldsymbol{d}_j(0)=\boldsymbol{d}_{0,j},
\end{aligned}
\end{equation}
where $\Pi_5$ projects onto the first five orbital elements $(a,\varepsilon,\iota,\Omega,\omega)$ and $\mathcal{J}_{\mathrm{CPA}}$ denotes the closest-point-of-approach penalty, evaluated separately over spacecraft--debris and spacecraft--spacecraft pairs with their respective safety radii (see Section~\ref{sec:methods_interactions}). This formulation expresses the desired collision-aware planning problem for a single scenario, but solving equation~\eqref{eq:instance_ocp} repeatedly would require expensive numerical optimization and would make supervised training dependent on a large library of precomputed optimal trajectories.

Instead, we train $\mathcal{G}_\theta$ as an amortized solution operator over a distribution of scenarios. At each training instance, we sample $(P_0,P_d,P_1,T)$, construct a debris set with an adversarial fraction, and generate trajectories using the biased baseline plus learned residual parameterization, namely
\begin{equation}
    \boldsymbol{x}_\theta(t;\boldsymbol{x}_0) = \bar{\boldsymbol{x}}(t) + \Delta\boldsymbol{x}_\theta(t;\boldsymbol{x}_0),
\end{equation}
where $\bar{\boldsymbol{x}}$ is the deterministic baseline of Section~\ref{sec:methods_bias_rollout} and $\Delta\boldsymbol{x}_\theta$ is the network residual, conditioned on the full scenario $(P_0,P_d,P_1,T)$. The rollout realizes the trajectory map of equation~\eqref{eq:operator_def} sample-wise: $\mathcal{G}_\theta(P_0,P_d,P_1,t,T)=\boldsymbol{x}_\theta(t;\cdot)_\#P_0$.
The network parameters are optimized by minimizing the expected self-supervised loss
\begin{equation}
\label{eq:amortized_training_loss}
\min_{\theta}\;
\mathbb{E}_{(P_0,P_d,P_1,T)\sim\mathcal{D}}
\left[
\lambda_f \mathcal{F}_\theta
+ \lambda_I \mathcal{I}_\theta
+ \lambda_T \mathcal{T}_\theta
+ \lambda_S \mathcal{S}_\theta
\right],
\end{equation}
where $\mathcal{D}$ is the training distribution over planning scenarios. 
The four terms are defined as follows:
\begin{equation}
    \mathcal{F}_\theta 
    = \mathbb{E}_{\boldsymbol{x}_0\sim P_0} \int_0^T
    \left\|
        \boldsymbol{g}(\boldsymbol{x}_\theta(t))^\dagger
        \left(
            \dot{\boldsymbol{x}}_\theta(t)-\boldsymbol{f}(\boldsymbol{x}_\theta(t))
        \right)
    \right\|_2^2\,dt
\end{equation}
is the fuel cost, the distributional form of equation~\eqref{equ:fuel_cost_raw};
and
\begin{equation}
    \mathcal{I}_\theta
    = \mathcal{J}_{\mathrm{CPA}}
    \left(
        \boldsymbol{x}_\theta(t;\cdot)_\# P_0,\; P_d
    \right)
\end{equation}
is the closest-point-of-approach interaction penalty of equation~\eqref{equ:J_cpa_def}, evaluated on the pushforward of $P_0$ under the rollout along the horizon;
and
\begin{equation}
    \mathcal{T}_\theta
    = \mathbb{E}_{(\boldsymbol{x}_0,\boldsymbol{x}_{\mathrm{des}})\sim(P_0,P_1)}
    \left\|
    \Pi_5\!\left(\boldsymbol{x}_\theta(T;\boldsymbol{x}_0)\right)
    -
    \Pi_5\!\left(\boldsymbol{x}_{\mathrm{des}}\right)
    \right\|_2^2
\end{equation}
is the terminal loss of equation~\eqref{eq:terminal_loss}, where the expectation is over paired samples, each spacecraft with its own assigned target;
and
\begin{equation}
    \mathcal{S}_\theta
    = \mathbb{E}_{\boldsymbol{x}_0\sim P_0} 
    \left\|
    \Pi_5\!\left(\boldsymbol{x}_\theta(0;\boldsymbol{x}_0)\right)
    -
    \Pi_5\!\left(\boldsymbol{x}_0\right)
    \right\|_2^2
\end{equation}
is the initial loss, which anchors the rollout at $\tau=0$ to the sampled initial state; in both anchoring terms the projection $\Pi_5$ excludes the true anomaly.
This objective trains the operator without ground-truth optimal trajectories or numerically generated trajectory labels, while preserving the structure of the per-instance optimal control problem in equation~\eqref{eq:instance_ocp}. The specific weights are given in Section~\ref{sec:training_setup}.

\subsection{Dynamic-feasibility finish via Gauss--Newton terminal targeting}\label{sec:methods_reprojection}

The operator's element-space rollout is not, by construction, the integral of a physical control sequence under exact two-body dynamics. We finish each rollout, per agent, with a single-shooting Gauss--Newton (GN) step with Levenberg--Marquardt damping. The control sequence $\boldsymbol{u}_{0:T-1}$ is the only decision variable; the state is the exact RK4 rollout $\boldsymbol{x}_{k+1}=\Phi^{\mathrm{RK4}}_{\Delta t}(\boldsymbol{x}_k,\boldsymbol{u}_k)$ from the fixed initial state, so every iterate is dynamically feasible and there are no dynamics constraints.

We target the orbit through its conserved vectors rather than its element angles: the specific angular momentum $\boldsymbol{h}=\boldsymbol{r}\times\boldsymbol{v}$ and the eccentricity vector $\boldsymbol{e}=(\boldsymbol{v}\times\boldsymbol{h})/\mu-\boldsymbol{r}/\|\boldsymbol{r}\|$. Both are invariant along an orbit, so the residual is phase-free, and smooth in $(\boldsymbol{r},\boldsymbol{v})$, so the Jacobian stays well-defined for the near-circular and near-equatorial orbits where element-angle residuals are singular. With targets $(\boldsymbol{h}^\star,\boldsymbol{e}^\star)$ from the goal orbit $P_1$, the residual and objective are
\begin{equation}
\label{eq:lm_residual}
\begin{gathered}
\boldsymbol{\rho}(\boldsymbol{u})=\big[(\boldsymbol{h}(T)-\boldsymbol{h}^\star)/s_h;\;(\boldsymbol{e}(T)-\boldsymbol{e}^\star)/s_e\big]\in\mathbb{R}^6,\\
\min_{\boldsymbol{u}}~\|\boldsymbol{\rho}(\boldsymbol{u})\|_2^2+\lambda_f\|\boldsymbol{u}\|_2^2 .
\end{gathered}
\end{equation}
Each iteration forms the Jacobian $\boldsymbol{J}=\partial\boldsymbol{\rho}/\partial\boldsymbol{u}\in\mathbb{R}^{6\times 3T}$ by automatic differentiation through the RK4 rollout, six vector--Jacobian products, one per residual component, sharing a single retained backward graph, and takes the damped Gauss--Newton step
\begin{equation}
\label{eq:lm_step}
\Delta\boldsymbol{u}=-\big(\boldsymbol{J}^\top\boldsymbol{J}+(\lambda_f+\mu)\boldsymbol{I}\big)^{-1}\big(\boldsymbol{J}^\top\boldsymbol{\rho}+\lambda_f\boldsymbol{u}\big),
\end{equation}
where $\mu\ge0$ is the Levenberg--Marquardt damping. The decision vector $\boldsymbol{u}$ has dimension $3T$ (hundreds to thousands), but the residual has only six components, so we never form the $3T\times3T$ system. Writing $\alpha=\lambda_f+\mu$ and $\boldsymbol{b}=-(\boldsymbol{J}^\top\boldsymbol{\rho}+\lambda_f\boldsymbol{u})$, the Woodbury identity collapses equation~\eqref{eq:lm_step} to a single $6\times6$ solve,
\begin{equation}
\label{eq:lm_woodbury}
\Delta\boldsymbol{u}=\tfrac{1}{\alpha}\Big(\boldsymbol{b}-\boldsymbol{J}^\top\big(\alpha\boldsymbol{I}_6+\boldsymbol{J}\boldsymbol{J}^\top\big)^{-1}\boldsymbol{J}\,\boldsymbol{b}\Big),
\end{equation}
whose cost is independent of the horizon length $T$ and identical for every agent. The swarm is therefore solved as one batched stack of $6\times6$ systems, the property that lets the finish run at $N=1000$, where a per-agent nonlinear program is intractable. The $6\times6$ inner solve is carried out in double precision to absorb the $1/\alpha$ cancellation when the fuel term is active ($\lambda_f>0$); for the pure terminal target ($\lambda_f=0$) the step reduces to the numerically benign minimum-norm form $\Delta\boldsymbol{u}=-\boldsymbol{J}^\top(\boldsymbol{J}\boldsymbol{J}^\top+\mu\boldsymbol{I}_6)^{-1}\boldsymbol{\rho}$.

\subsection{Training setup}\label{sec:training_setup}

Each training step draws one LEO scenario. The number of spacecraft $N \in \{1,\dots,10\}$ is sampled uniformly; a cluster-center orbit is drawn from standard LEO element bounds (semi-major axis $6848$--$8748$~km, eccentricity ${\leq}0.1$, unrestricted angles, periapsis above Earth $+100$~km), and the agents are placed within a $50$~m ball of the center in both position and matched velocity, which makes agent--agent conflict unavoidable at every $N \geq 2$. Targets are per-agent but converging: one deviation vector at exactly the $1\%$ or $10\%$ class magnitude (random sign per element, true anomaly free) is applied to every agent's own initial elements, so an unmodified rollout stays in conflict to arrival. The horizon is drawn from $[1,12]$~h, and debris consist of one adversarial crossing-orbit object per agent ($M=N$; Section~\ref{sec:methods_interactions}). The loss weights are $(\lambda_f,\lambda_I,\lambda_T,\lambda_S) = (10^{-2},10^{3},10^{2},10^{2})$ with hinge scale $\kappa=10^4$, and the anchoring losses are normalized element-wise by the scale vector $[1.37, 0.1, \pi, 2\pi, 2\pi]$. The network (hidden width $1024$, five cross-attention layers, eight heads, ${\approx}48$ million parameters) is trained with Adam at learning rate $2\times10^{-5}$ under a reduce-on-plateau schedule, gradient-norm clipping at $100$, and single-precision arithmetic for $14{,}000$ iterations; non-finite guards and a loss-spike filter protect against the stiff dynamics. At inference the operator is queried on a uniform grid with a constant $120$~s physical timestep in a single batched forward pass, with the true anomaly integrated sequentially from its predicted rate correction.

\section{Experimental Results}\label{sec:results}

We evaluate whether the learned neural operator can (i) generate low-cost orbital transfers, (ii) avoid close approaches in dense debris fields, and (iii) generalize beyond the training distribution in both swarm size and debris density. The operator is trained on short-duration missions of 1--12 hours, with agent counts $N\in\{1,\dots,10\}$ and one adversarially placed debris object per agent. We report both interpolation performance within this regime and extrapolation to swarms of up to $N=1000$ agents amid the full ${>}11{,}000$-object catalog.

We evaluate the planner over a $2\times2$ family of test conditions in which every scenario draws its $N$ spacecraft initial states from the real Two-Line Element (TLE) catalog. The \emph{maneuver} axis sets the retargeting magnitude: a \emph{minor} maneuver perturbs each orbital element by exactly $1\%$ of its scale (random sign per element; station-keeping), while a \emph{major} maneuver perturbs each element by exactly $10\%$ (rapid response). The \emph{debris} axis sets the threat construction. A \emph{debris} scenario surrounds the swarm with ambient catalog objects, the routine collision-avoidance regime, whereas an \emph{adversarial} scenario places one worst-case object on each method's own debris-unaware predicted path, so a planner that does not condition on the debris field is struck unless it actively deviates. Proximity is reported as a per-spacecraft rate, the percentage of planned maneuvers that pass within $100$~m of another agent or debris object, and all reported performance values are medians over $500$ Monte Carlo trials per cell.

\subsection{Adversarial and debris-field trajectory planning}

Table~\ref{tab:main_performance} reports terminal accuracy, fuel cost, and per-spacecraft proximity across the four scenarios and swarm sizes. We compare the operator-warm Gauss--Newton finish (GNw), which closes each operator rollout onto exact two-body dynamics, with the same finish cold-started without the operator seed (GNc). GNc reaches the same target orbit but lacks the operator's learned collision avoidance, so the GNw--GNc gap isolates what the operator contributes. Both run batched across the swarm at every size, including $N=1000$, where a per-agent nonlinear program is intractable.

\begin{table}[!t]
\centering
\setlength{\tabcolsep}{2.8pt}\footnotesize
\caption{Terminal accuracy, fuel cost, and collision behavior across debris and adversarial environments}
\label{tab:main_performance}
\begin{tabular}{@{}llcccccc@{}}
\toprule
 & & \multicolumn{2}{c}{Error (\%)} & \multicolumn{2}{c}{$\Delta v$ (km/s)} & \multicolumn{2}{c}{Prox$_{100}$ (\%)} \\
\cmidrule(lr){3-4}\cmidrule(lr){5-6}\cmidrule(lr){7-8}
Scenario & $N$ & GNw & GNc & GNw & GNc & GNw & GNc \\
\midrule
Debris, minor & 1    & 0.0007 & 0.0007 & 0.585 & 0.581 & 0    & 2.60 \\
 & 10   & 0.0007 & 0.0007 & 0.590 & 0.583 & 0    & 3.48 \\
 & 100  & 0.0007 & 0.0006 & 0.590 & 0.582 & 0.09 & 3.45 \\
 & 1000 & 0.0007 & 0.0006 & 0.589 & 0.582 & 0.28 & 3.50 \\
\midrule
Debris, major & 1    & 0.0186 & 0.0140 & 6.004 & 5.592 & 0    & 2.00 \\
 & 10   & 0.0206 & 0.0148 & 6.061 & 5.733 & 0    & 3.42 \\
 & 100  & 0.0197 & 0.0148 & 6.084 & 5.826 & 0.05 & 3.35 \\
 & 1000 & 0.0197 & 0.0146 & 6.104 & 5.815 & 0.11 & 3.30 \\
\midrule
Adversarial, minor & 1    & 0.0007 & 0.0007 & 0.592 & 0.583 & 0    & 99.8 \\
 & 10   & 0.0007 & 0.0007 & 0.589 & 0.582 & 0    & 99.5 \\
 & 100  & 0.0007 & 0.0006 & 0.589 & 0.582 & 0.03 & 99.6 \\
 & 1000 & 0.0007 & 0.0006 & 0.589 & 0.582 & 0.21 & 99.6 \\
\midrule
Adversarial, major & 1    & 0.0210 & 0.0176 & 6.101 & 5.633 & 0    & 99.6 \\
 & 10   & 0.0213 & 0.0148 & 6.060 & 5.735 & 0    & 99.3 \\
 & 100  & 0.0197 & 0.0139 & 6.082 & 5.827 & 0.01 & 99.5 \\
 & 1000 & 0.0199 & 0.0151 & 6.090 & 5.873 & 0.07 & 99.5 \\
\bottomrule
\end{tabular}
\end{table}

The debris scenarios are the realistic operating regime: real spacecraft initial conditions together with the actual catalogued debris field, the setting a swarm would face in today's crowded low Earth orbit. Here the operator-warm finish keeps close approaches rare, at or below $0.28\%$ of maneuvers within $100$~m, while the debris-blind GNc approaches within $100$~m on $2$--$3.5\%$ of maneuvers, an order of magnitude more, at a fuel cost within ten percent of GNw's. This behavior holds far beyond the $N\leq10$ training range, degrading gradually rather than abruptly out to $N=1000$, with terminal error held at $10^{-4}$--$10^{-2}\%$ throughout.

The same learned avoidance extends to an adversarial setting. An adversarial object is one positioned directly on a spacecraft's intended path, so that failing to deviate means near-certain collision. We construct this worst case by seeding one such object on each method's own debris-unaware path. A planner blind to it is struck almost every time (GNc, $99.3$--$99.8\%$ of maneuvers at all sizes), whereas the operator-warm finish clears it on essentially every maneuver (at most $0.21\%$ at $N=1000$, essentially none of which involves the threat itself) at comparable fuel and accuracy. Nominal debris avoidance and defensive evasion are therefore one capability, driven by the same conditioning on the surrounding object field and exercised here against a deliberately harder threat.

In the single-agent setting ($N=1$) we additionally solve the full optimal-control problem with IPOPT\cite{wachter2006ipopt} as a reference: a nonlinear program over the controlled two-body dynamics that minimizes fuel while driving the agent to its target orbit and enforcing hard debris-avoidance constraints. It is tractable only when both the agent and debris counts are small, so we report it at $N=1$ in the adversarial scenarios, where each agent faces a single worst-case object; against the full catalog it is intractable even at $N=1$, as it imposes a separate collision constraint at every debris object and time step. On the real single-agent transfers it attains $0.0012\%$ terminal error at $\Delta v = 0.582$~km/s for the minor maneuver and $0.045\%$ at $5.52$~km/s for the major maneuver. The operator-warm finish GNw approaches this optimum, reaching comparable terminal accuracy (Table~\ref{tab:main_performance}) at a fuel cost within $2\%$ on the minor maneuver and within $11\%$ on the major maneuver, while remaining batched and scalable to the swarm sizes where IPOPT cannot run.

IPOPT is thus a quality benchmark rather than a scalable baseline: although the operator is trained only on $N\leq10$, it retains bounded terminal errors and low proximity rates at $N=100$ and $N=1000$ (Table~\ref{tab:main_performance}), where a per-agent nonlinear program is no longer practical as a routine online planner.

\begin{figure}[!t]
\centering
\includegraphics[width=0.49\columnwidth]{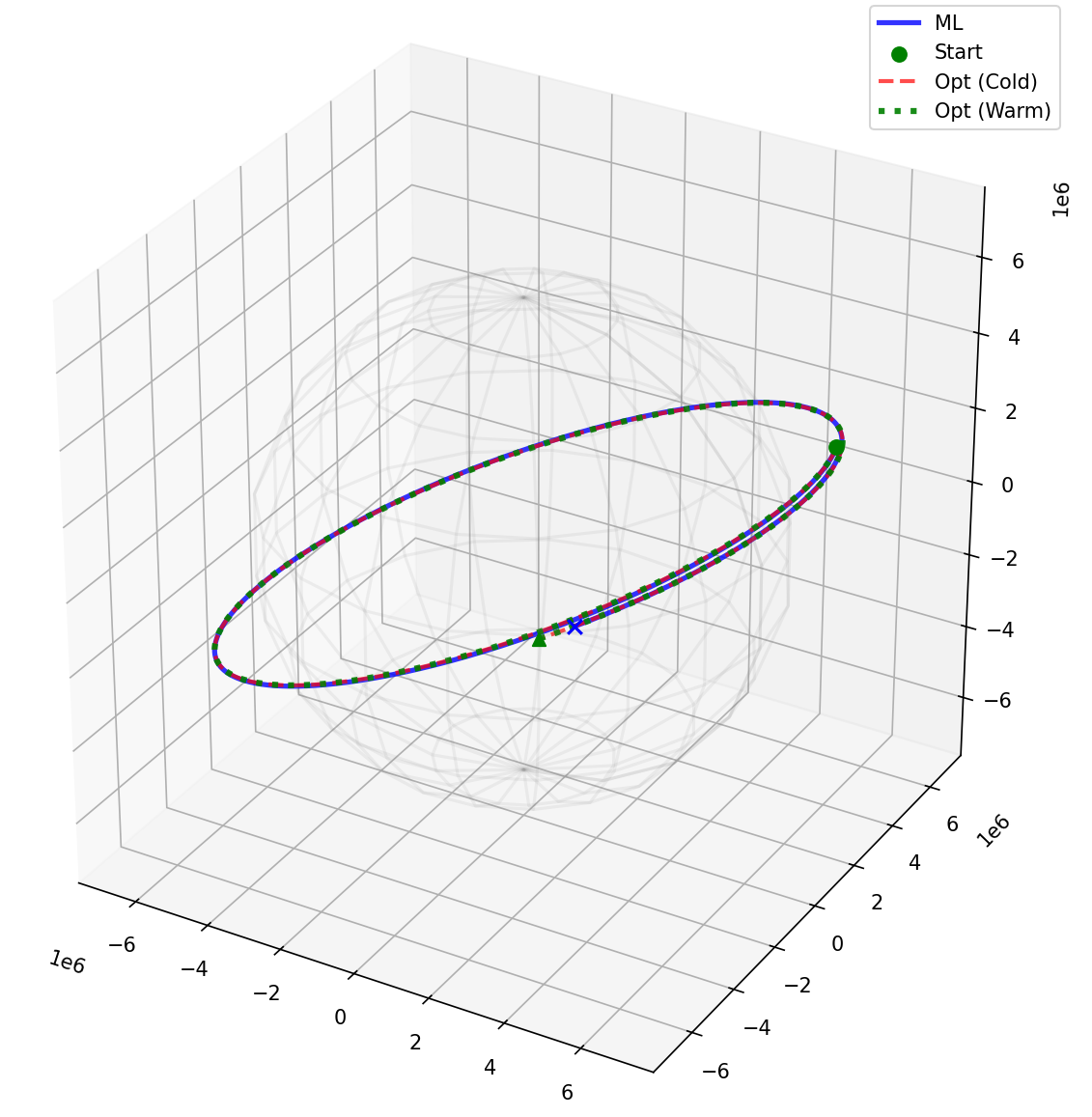}\hfill
\includegraphics[width=0.49\columnwidth]{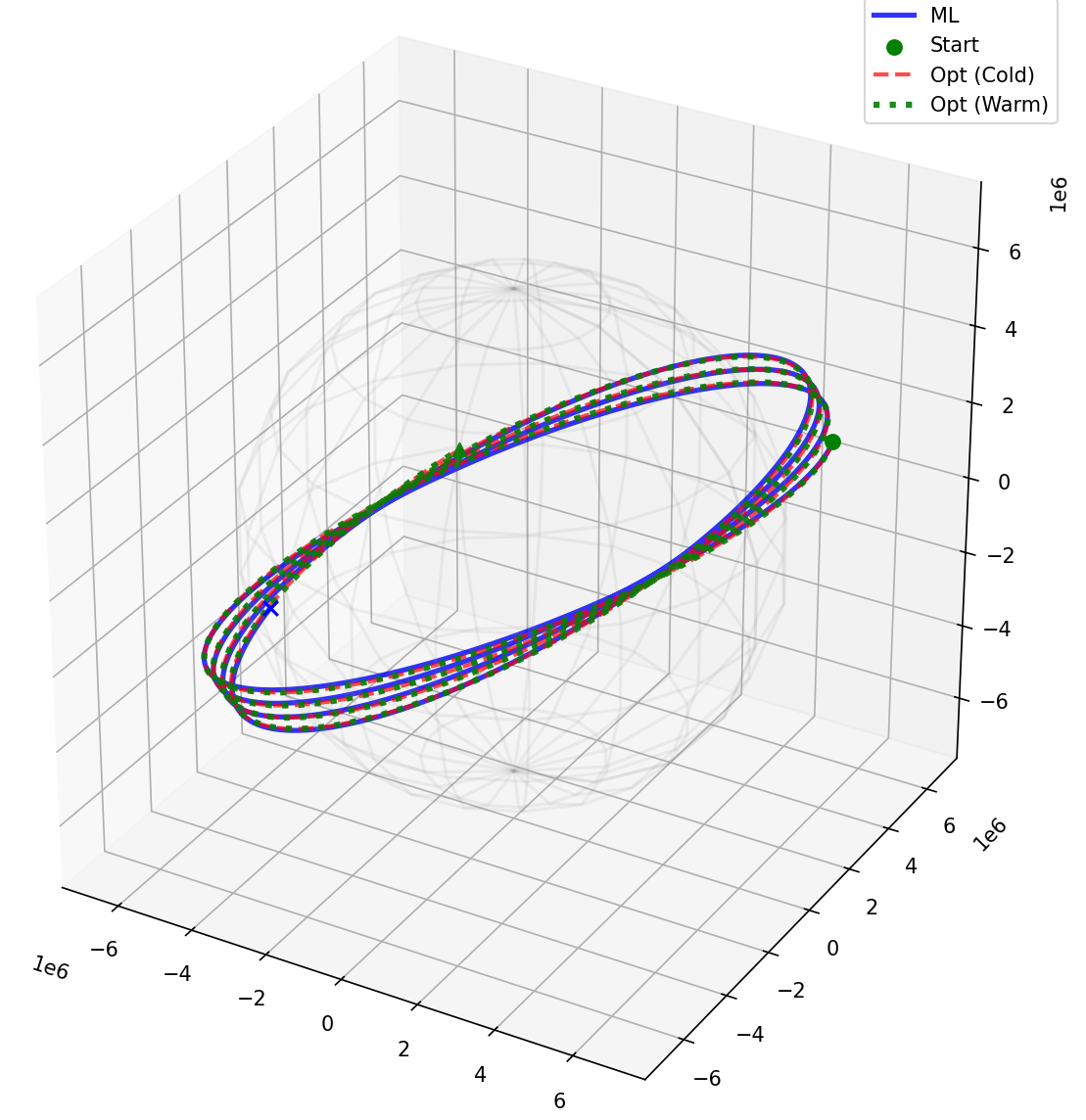}\\[2pt]
\includegraphics[width=0.49\columnwidth]{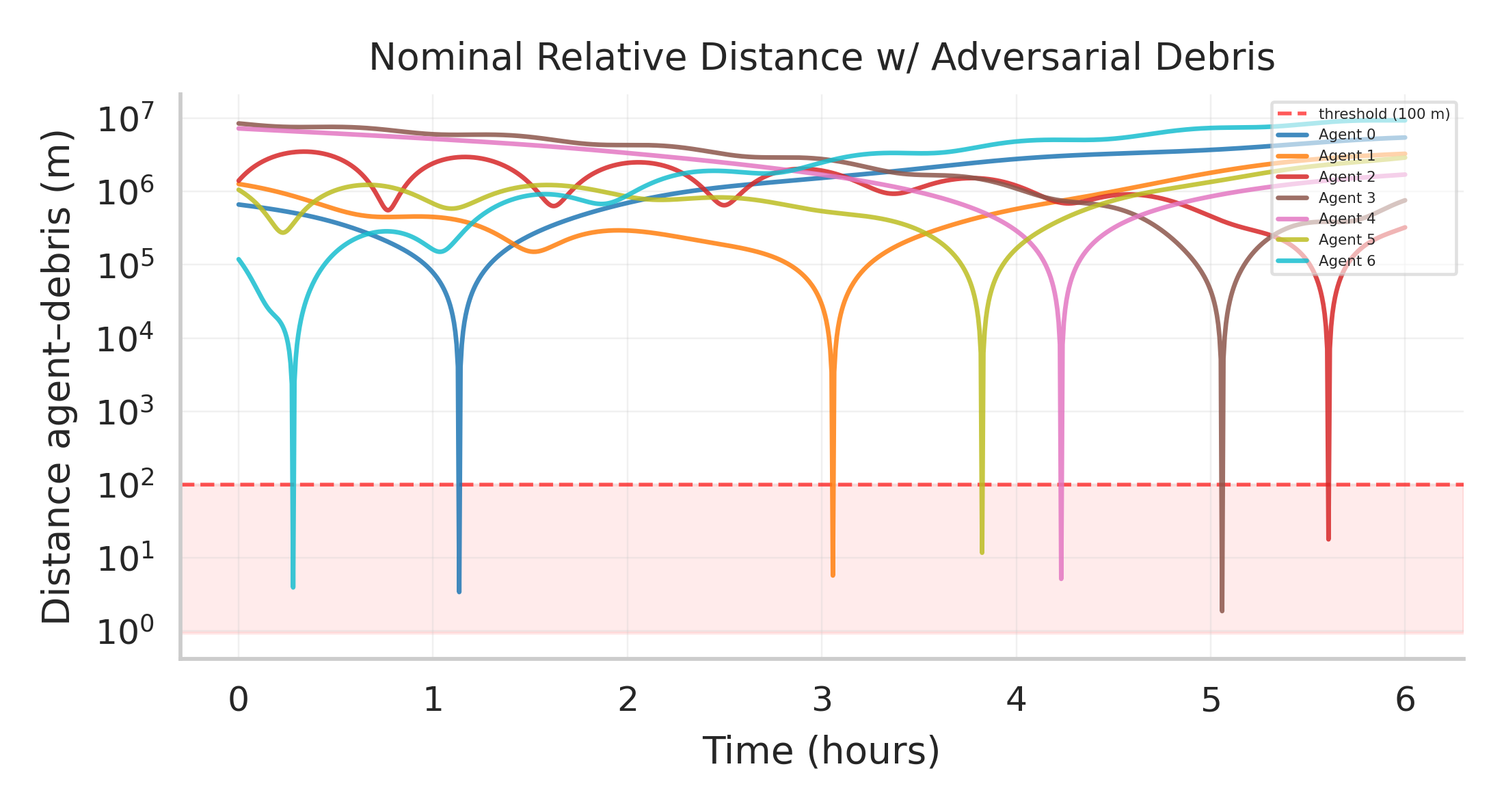}\hfill
\includegraphics[width=0.49\columnwidth]{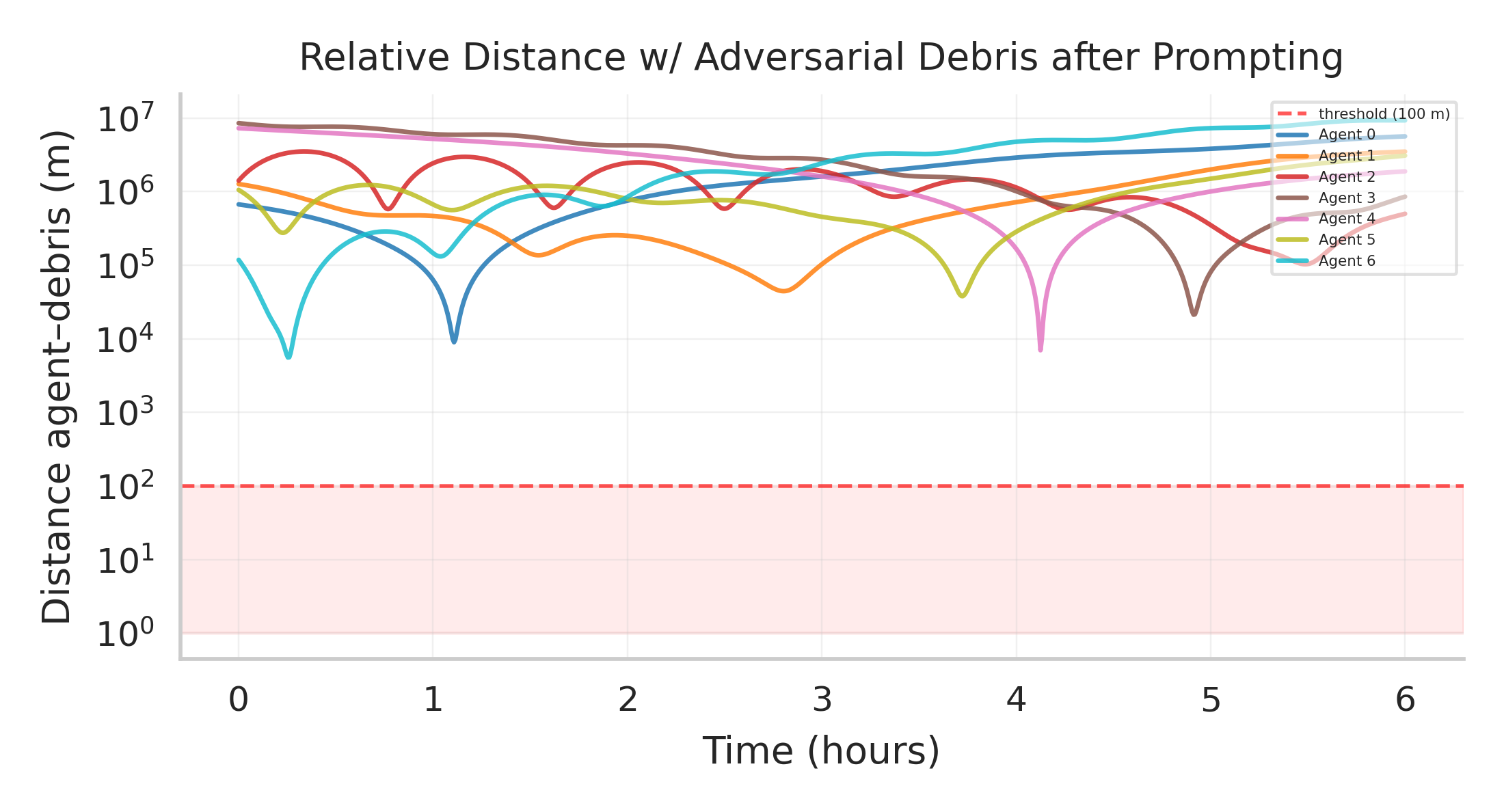}\\[2pt]
\includegraphics[width=0.49\columnwidth]{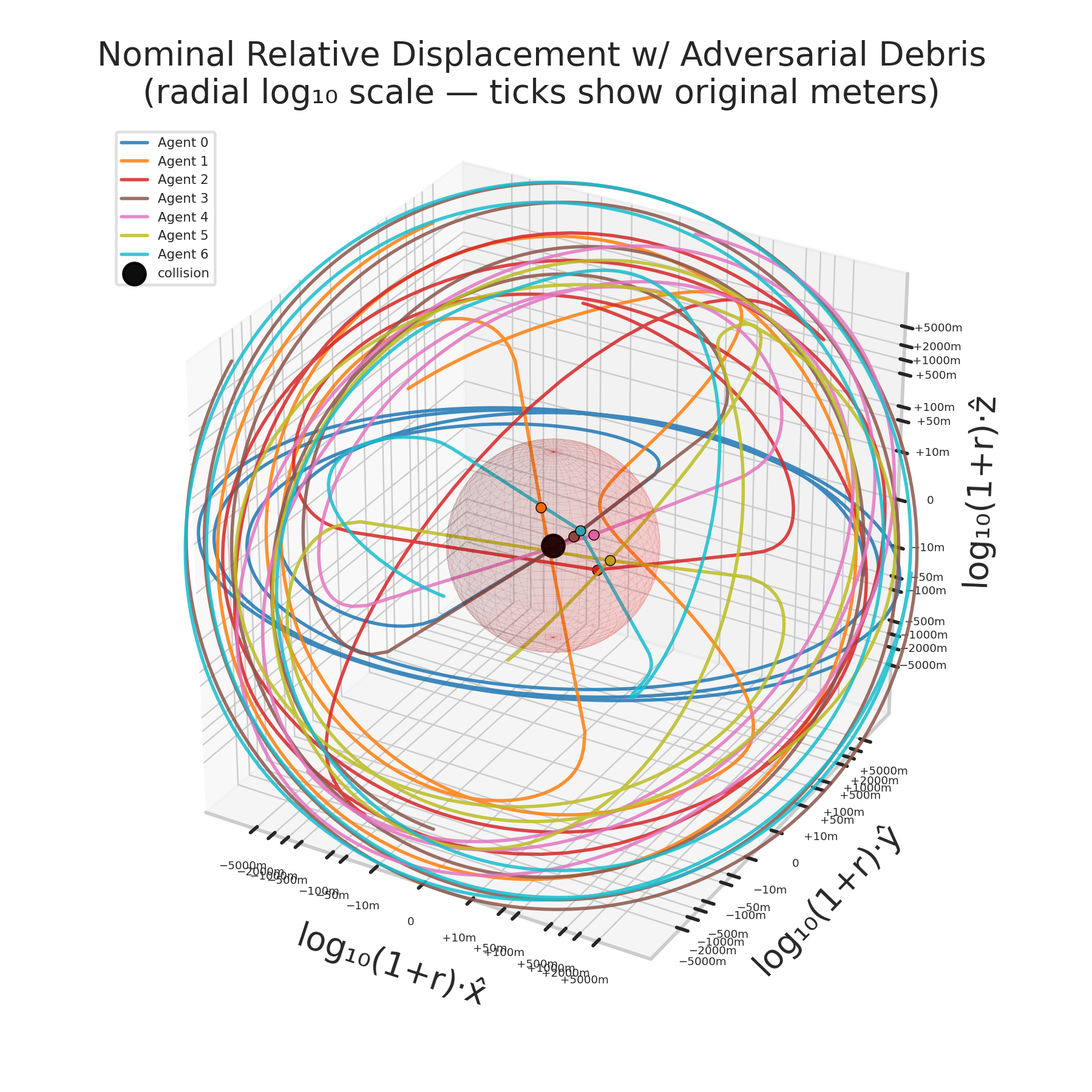}\hfill
\includegraphics[width=0.49\columnwidth]{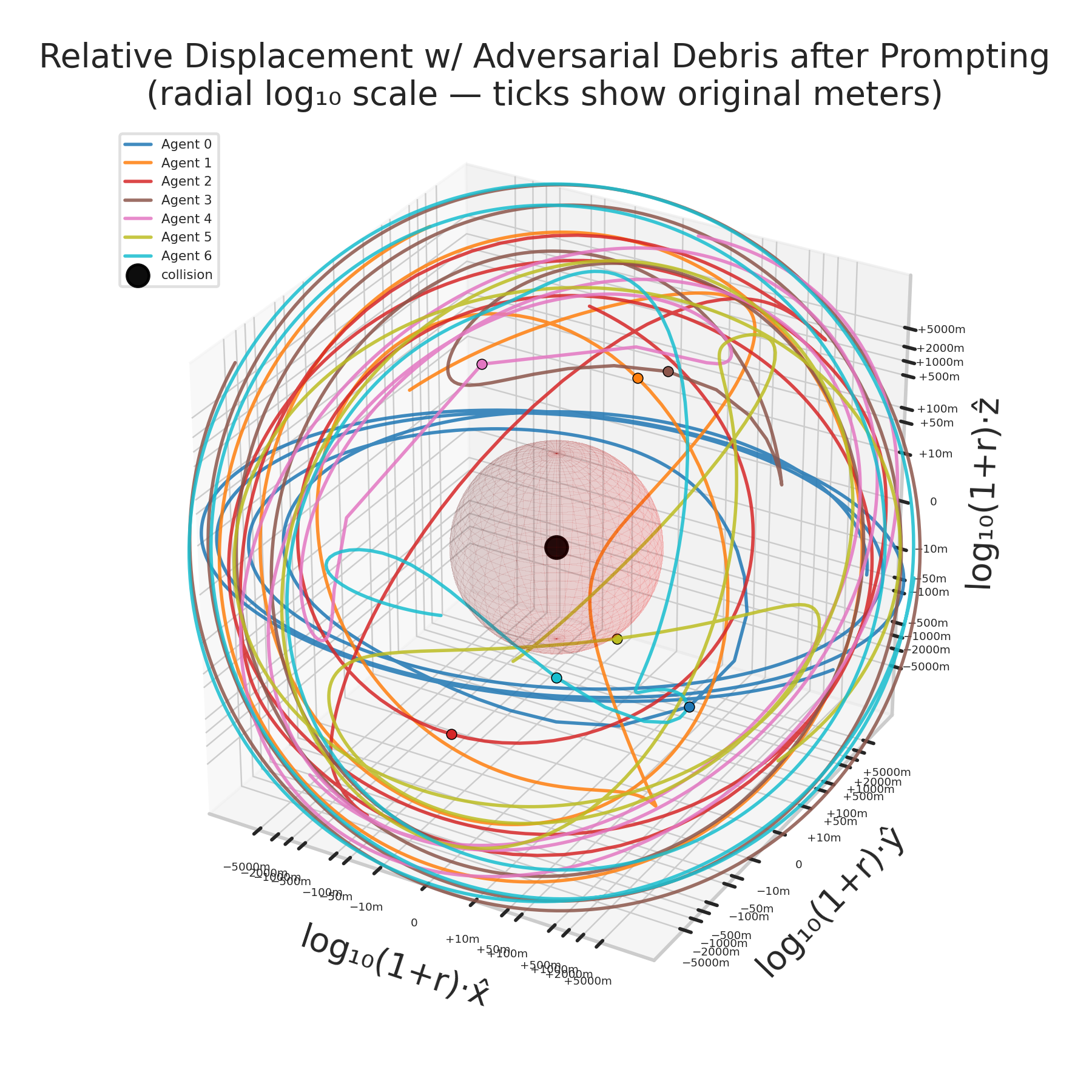}

\caption{Learned transfer geometry and debris-conditioned avoidance. Top: learned single-agent transfers match nonlinear optimal-control solutions for minor and major maneuvers. Middle: omitting debris information produces predicted close approaches, whereas debris conditioning keeps separation above the $100$~m threshold. Bottom: log-scaled relative trajectories show the nominal rollout entering the debris-centered danger region and the conditioned trajectory clearing it.}
\label{fig:qualitative_results}
\end{figure}

Figure~\ref{fig:qualitative_results} compares learned single-agent transfers with nonlinear optimal-control solutions and illustrates the effect of debris conditioning in an adversarial setting. The learned trajectories recover the optimized transfer geometry for both minor and major maneuvers. When adversarial debris is provided to the operator, the predicted trajectory changes to preserve separation above the $100$~m threshold. When the same debris information is omitted, multiple close approaches are predicted. The debris input thus directly shapes the trajectory toward avoidance.

\begin{figure}[!t]
\centering
\includegraphics[width=\columnwidth]{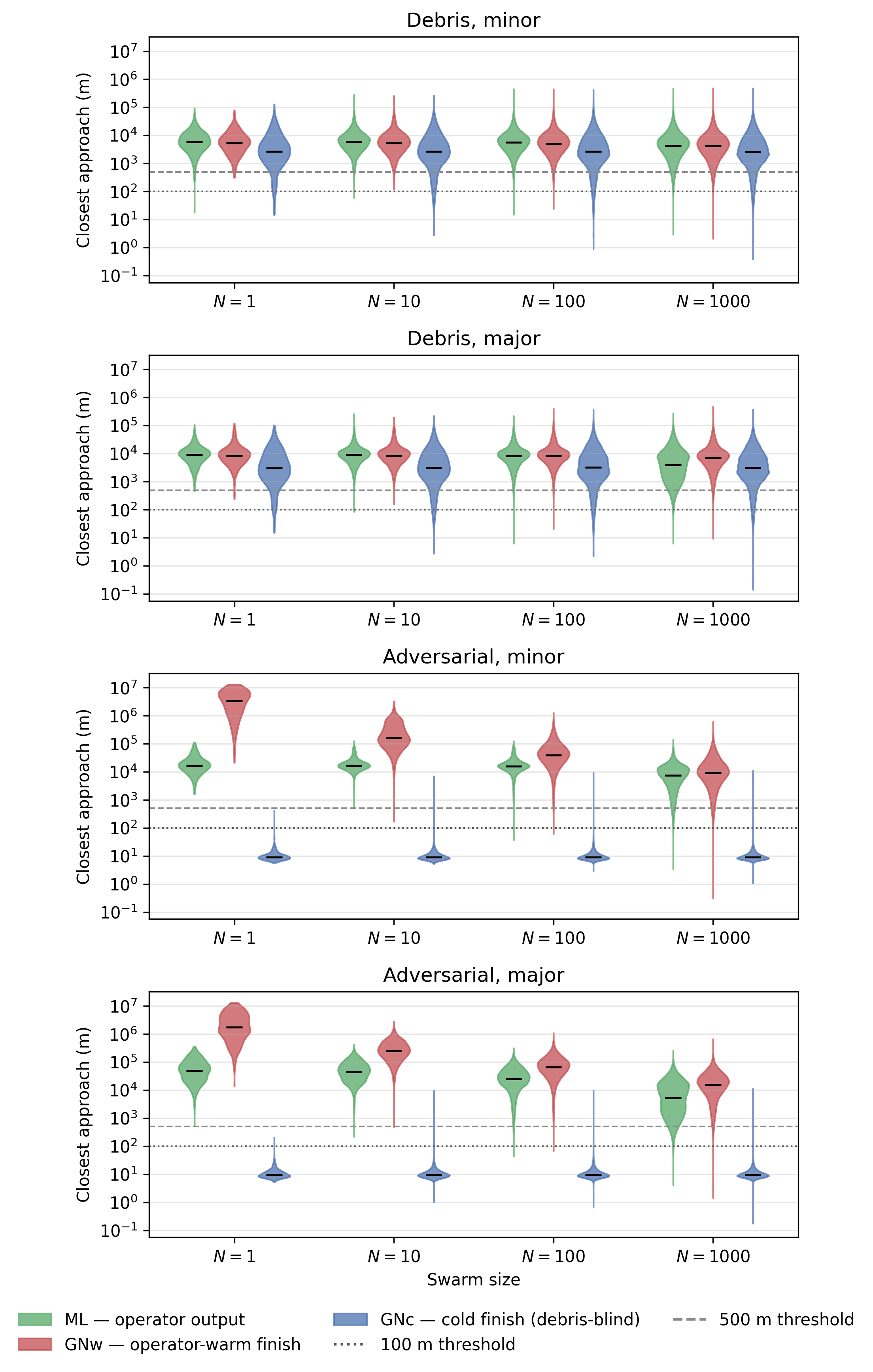}
\caption{Distribution of each spacecraft's closest approach ($500$ trials per cell): operator output (ML, green), operator-warm finish (GNw, red), and cold debris-blind finish (GNc, blue). Black bars mark medians; dotted and dashed lines the $100$ and $500$~m thresholds.}
\label{fig:minsep_violins}
\end{figure}

Although the operator is trained only on $N\leq10$, the finished terminal error remains tightly bounded when extrapolating to $N=100$ and $N=1000$ across all four scenarios (Table~\ref{tab:main_performance}), indicating stable degradation rather than abrupt failure far beyond the training distribution. All scenarios draw their initial conditions, and the debris scenarios their debris fields, from a public Space-Track Two-Line Element (TLE) catalog snapshot containing over $11{,}000$ resident space objects; each TLE is propagated to its epoch with the SGP4 model~\cite{vallado2006sgp4} and converted to Cartesian states.

\subsection{Dynamic feasibility via a Gauss--Newton finish}
\label{sec:results_reprojection}

We close each rollout with a  single-shooting step over the control sequence, a phase-free, non-singular terminal target, under a fuel regularizer. Warm-started from the operator (GNw) the finish inherits the operator's collision-aware geometry; cold-started (GNc) it reaches the same target orbit without that geometry. The full construction, including the conserved-vector residual and its batched solution, is derived in Section~\ref{sec:methods}.

GNw drives terminal error to $10^{-3}$--$10^{-2}\%$ (Table~\ref{tab:main_performance}), one to two orders below the operator's raw element-space output and comparable to the single-agent optimum, at a fuel cost close to that single-agent optimum (Table~\ref{tab:main_performance}), and the accuracy is bounded across three orders of magnitude in $N$.

\subsection{Collision avoidance}
\label{sec:results_collision}

Table~\ref{tab:main_performance} reports medians; Fig.~\ref{fig:minsep_violins} shows the full distribution of each spacecraft's closest approach for the operator's raw output (ML), the operator-warm finish (GNw), and the debris-blind cold finish (GNc). Against the adversarial threat, GNc is struck on almost every maneuver ($99.3$--$99.8\%$) and its distribution collapses onto the threat at ${\sim}10$~m, whereas the operator clears it essentially always (at most $0.21\%$), with the ML and GNw mass sitting kilometers above the $100$~m threshold; under catalog debris, GNc approaches a debris object an order of magnitude more often than GNw. Throughout, GNw tracks ML to within a few tenths of a percent: the dynamics-closing finish preserves the learned avoidance rather than eroding it.

Splitting the residual by pair type shows that avoidance of \emph{external} objects is effectively complete: GNw's agent--debris rate is at most $0.002\%$ in the adversarial scenarios and at or below $0.06\%$ at $100$~m under catalog debris, against $3.2\%$ for GNc. What remains is almost entirely agent--agent, appears only once the swarm is dense ($N\geq100$), and is directly shaped by training: the deliberately conflicting scenario construction (Section~\ref{sec:training_setup}) reduces swarm-internal proximity three- to eight-fold relative to the avoidance-free GNc ($0.07$--$0.22\%$ versus $0.51$--$0.67\%$ at $N=1000$); further improving agent--agent deconfliction is a direction for future work.

\subsection{Grid-independent proximity scoring}
\label{sec:results_cpa_ablation}

\begin{figure}[!t]
\centering
\includegraphics[width=\linewidth]{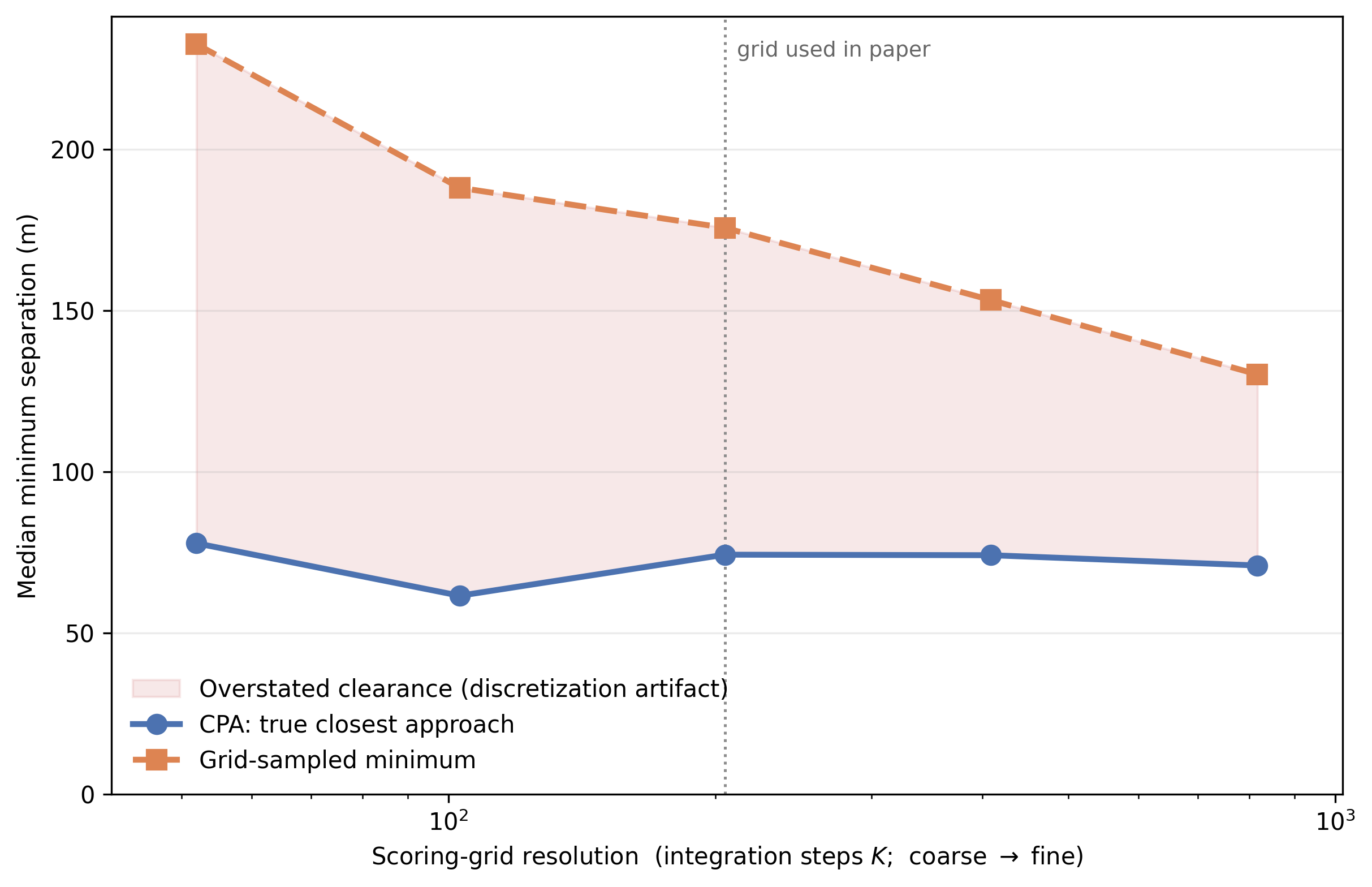}
\caption{Grid-independence of CPA scoring for the densest debris cell ($N=1000$, minor): grid-sampled minimum separation (orange, dashed) versus within-interval CPA measure (blue, solid). The shaded band is clearance that does not physically exist; the vertical line marks the grid used in this paper.}
\label{fig:cpa_ablation}
\end{figure}

Every separation reported in this paper is measured with the closest-point-of-approach refinement of equation~\eqref{eq:cpa_linear_within_interval}: within each rollout interval we solve analytically for the instant of minimum separation rather than reading off the smallest grid-point distance. Fig.~\ref{fig:cpa_ablation} shows why this matters. Re-scoring the densest cell while sweeping the grid from coarse to fine, the grid-sampled minimum overstates the true clearance by roughly $1.5$--$2\times$ and drifts with resolution, whereas the CPA measure is flat across the sweep, recovering the same physical closest-approach distance at every resolution. The reported proximity rates are therefore faithful and grid-independent, and the same behavior holds across all scenarios and swarm sizes.

\subsection{Runtime and scaling}
\label{sec:results_runtime}

\begin{figure}[!t]
\centering
\includegraphics[width=\columnwidth]{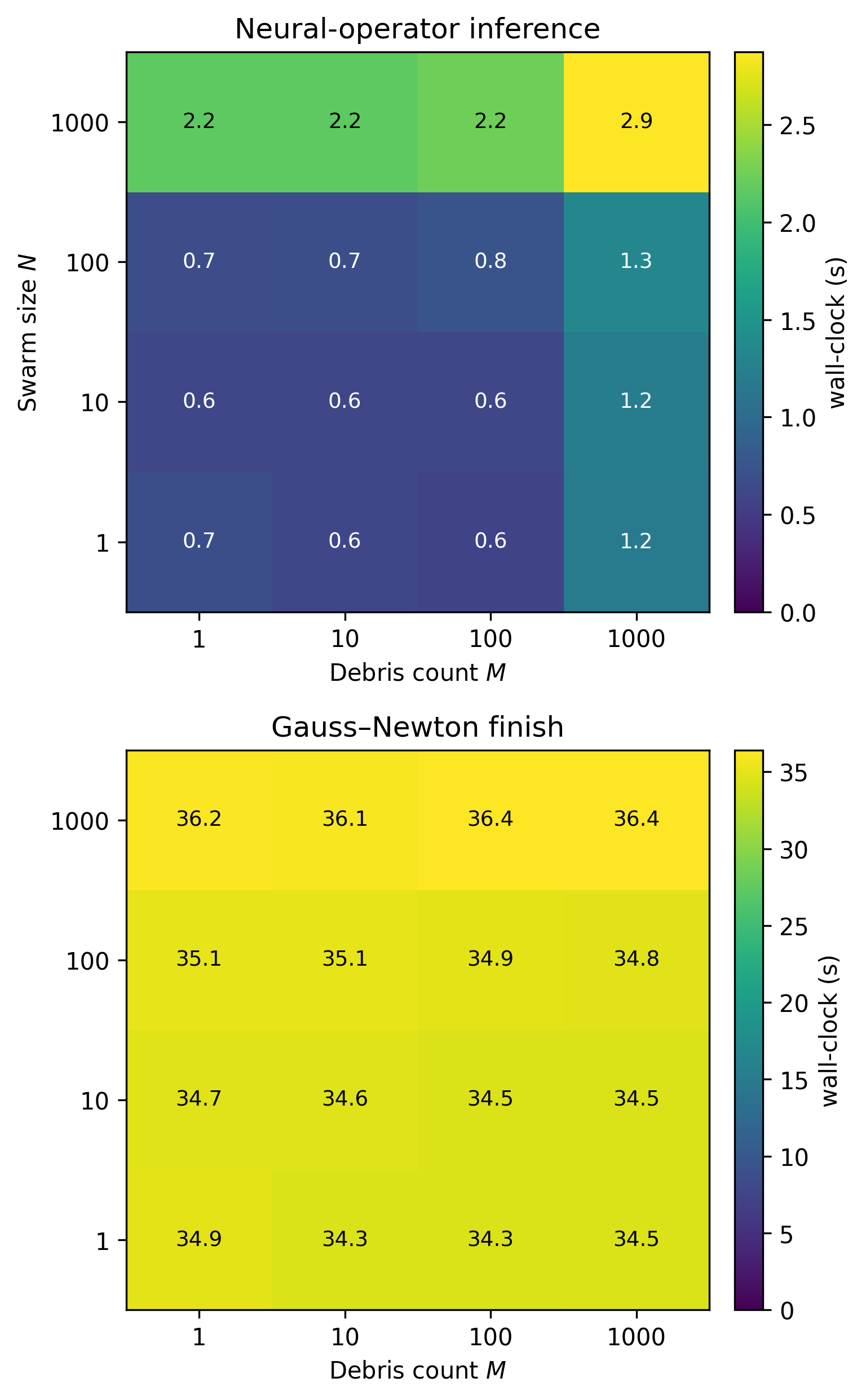}
\caption{Median wall-clock time (s) on one NVIDIA RTX 2080 Ti versus swarm size $N$ and debris count $M$ at a 6-hour horizon: neural-operator inference (top) and Gauss--Newton finish (bottom). Both panels use zero-based color scales.}
\label{fig:runtime_heatmap}
\end{figure}

We measured wall-clock time for both stages on the same workstation used for the Monte Carlo evaluations (a single NVIDIA RTX 2080 Ti, 11~GB). Both the operator rollout and the Gauss--Newton finish use a constant physical timestep $\Delta t = 120$~s, so the number of integration steps scales with mission duration ($K \approx T/\Delta t$, from ${\approx}31$ for a 1-hour transfer to ${\approx}361$ for the full 12-hour horizon) rather than with swarm size. Figure~\ref{fig:runtime_heatmap} reports the median time per $(N,M)$ cell at a representative 6-hour horizon ($K\approx181$).

Operator inference stays below 3~seconds throughout, rising only from $0.7$~s at $N=1$ to $2.9$~s at $N=1000$ with $M=1000$ debris. The Gauss--Newton finish reduces to a batched, fixed-size linear solve per agent (Section~\ref{sec:methods}), so its cost is set by the sequential RK4 rollouts in each iteration rather than by $N$ or $M$: it runs in ${\approx}35$~s and is essentially flat in swarm size ($34.9$~s at $N=1$, $36.4$~s at $N=1000$). The finish dominates, so the full pipeline replans the entire swarm in under a minute at the 6-hour horizon and roughly twice that at 12 hours; neither stage scales combinatorially with agent or debris count.

\subsection{Duration generalization}

The operator is trained on mission durations up to 12 hours and is not expected to extrapolate reliably to substantially longer horizons in a single rollout. Its inference latency suggests embedding it in a closed-loop receding-horizon controller that re-queries with updated spacecraft and debris states while remaining within the training distribution's temporal support. We do not evaluate this mode here, and all reported metrics are single-shot rollouts on the trained horizon.

\section{Discussion}\label{sec:discussion}

This work demonstrated that collision-aware trajectory planning for an entire spacecraft swarm can be amortized into a single forward pass of a permutation-equivariant neural operator, trained without optimal-trajectory labels from self-supervised physics objectives and adversarial threats generated against the model's own rollouts. Where classical planners re-solve a nonlinear program per agent and per scenario, the operator absorbs that cost at training time. Trained on ten spacecraft, it transfers zero-shot to swarms two orders of magnitude larger amid the full catalogued debris field.

Several limitations remain. For very small maneuvers the finished $\Delta v$ is mildly suboptimal, because the smooth quadratic control surrogate biases the warm start away from the sharp impulse-like profiles that minimize fuel there. Accuracy also degrades outside the trained 12-hour horizon. The interaction penalties are soft and the finish carries no collision term, so the method offers no worst-case collision-avoidance guarantee. At the largest swarms the residual proximity is almost entirely agent--agent (Table~\ref{tab:main_performance}; Section~\ref{sec:results_collision}). Both training and evaluation assume deterministic two-body Keplerian motion, without $J_2$, atmospheric drag, solar-radiation pressure, third-body perturbations, state-estimation uncertainty or thrust execution error, which matter operationally at the $100$--$500$~m thresholds considered here. We plan to address these limitations in future work.

As orbits grow more congested, planning methods whose cost scales with a single batched inference rather than with the number of pairwise constraints will become a prerequisite for swarm autonomy. The recipe demonstrated here, self-supervised physics losses, adversarial scenario generation and a certified numerical finish, offers a template for scalable, collision-aware multi-agent planning wherever swarms move through contested, cluttered environments.

\section*{Acknowledgment}
This research received no external funding. Large language model tools (Anthropic Claude models, Claude Opus 4.8 and Claude Fable 5) assisted with code development and manuscript editing; all methods, results, analyses, and conclusions were developed and verified by the authors, and no generative artificial intelligence was used to create any figure or image. The Two-Line Element catalog is publicly available from Space-Track (\url{https://www.space-track.org}; snapshot of March 24, 2025); data, trained weights, evaluation outputs, and code will be released upon publication.

\bibliographystyle{IEEEtaes}
\bibliography{references}

\end{document}